\documentclass[conference]{IEEEtran}
\IEEEoverridecommandlockouts
\usepackage{cite}
\usepackage{amsmath,amssymb,amsfonts}
\usepackage{algorithmic}
\usepackage{graphicx}
\usepackage{textcomp}
\usepackage{xcolor}
\usepackage{float}
\usepackage{mathtools}
\usepackage{subfigure}
\usepackage{soul}
\usepackage{multirow,color,todonotes}
\usepackage{array} 
\usepackage{tikz,balance}
\usepackage{url}
\usepackage{enumitem,xcolor,booktabs,colortbl}
\usepackage[linesnumbered, ruled]{algorithm2e}
\SetKwRepeat{Do}{do}{while}
\def\BibTeX{{\rm B\kern-.05em{\sc i\kern-.025em b}\kern-.08em
    T\kern-.1667em\lower.7ex\hbox{E}\kern-.125emX}}

\usepackage{xspace}
\newcommand{\wtm}{\textit{Word2Minecraft}\xspace}

\begin{document}

\title{Word2Minecraft: Generating 3D Game Levels through Large Language Models} 


\author{
    \IEEEauthorblockN{
        Shuo Huang\IEEEauthorrefmark{1}, 
        Muhammad Umair Nasir\IEEEauthorrefmark{1}\IEEEauthorrefmark{2}, 
        Steven James\IEEEauthorrefmark{2}, 
        Julian Togelius\IEEEauthorrefmark{1}
    }
    \IEEEauthorblockA{\IEEEauthorrefmark{1}%
        \textit{Department of Computer Science and Engineering}\\
        \textit{Tandon School of Engineering, New York University, New York 10012, United States}}
    \IEEEauthorblockA{\IEEEauthorrefmark{2}%
        \textit{School of Computer Science and Applied Mathematics}\\
        \textit{University of the Witwatersrand, South Africa}}
}
 
\IEEEoverridecommandlockouts

\IEEEpubid{
\hspace{\columnsep}\makebox[\columnwidth]{ }}

\maketitle

\IEEEpubidadjcol

\begin{abstract}
We present \wtm, a system that leverages large language models to generate playable game levels in Minecraft based on structured stories. The system transforms narrative elements---such as protagonist goals, antagonist challenges, and environmental settings---into game levels with both spatial and gameplay constraints. We introduce a flexible framework that allows for the customization of story complexity, enabling dynamic level generation. The system employs a scaling algorithm to maintain spatial consistency while adapting key game elements. We evaluate \wtm using both metric-based and human-based methods. Our results show that GPT-4-Turbo outperforms GPT-4o-Mini in most areas, including story coherence and objective enjoyment, while the latter excels in aesthetic appeal. We also demonstrate the system's ability to generate levels with high map enjoyment, offering a promising step forward in the intersection of story generation and game design. We open-source the code at https://github.com/JMZ-kk/Word2Minecraft/tree/word2mc\_v0

\end{abstract}


\section{Introduction}

Procedural content generation (PCG) is a widely used technique in game development that reduces the manual effort required for content creation while providing dynamic and diverse gameplay experiences. Story-driven level generation \cite{jenkins2004game}, a subfield of PCG, focuses on designing game levels where the narrative plays a central role in shaping their structure \cite{cook2016pcg}. However, traditional PCG methods for generating such levels typically rely on predefined rule-based algorithms, which often struggle to produce coherent and contextually meaningful game environments \cite{dahren2021usage,hartsook2011toward}.

Recent advances in Large language models (LLMs) have introduced new approaches to game content generation \cite{maleki2024procedural}. Given their strong ability to understand and produce human-like text \cite{wang2024guiding}, LLMs are well-suited for story generation. Game developers can leverage these models to create diverse narratives and design levels that align with them, highlighting a clear opportunity to bridge story generation with procedural level design.

Minecraft~\cite{duncan2011minecraft}, a widely popular 3D sandbox game, serves as an ideal platform for procedural level generation. However, translating stories into structured and playable Minecraft levels remains a challenging problem, requiring an intelligent mapping between language and level design.
To address this challenge, we present \wtm, an LLM-based system for generating Minecraft levels from structured stories---sequences of events constrained by spatial and gameplay requirements. The system first produces narratives and then translates them into playable game levels. To ensure spatial consistency, \wtm incorporates a scaling algorithm that dynamically adjusts the size of key game elements.

Both metric-based methods and human-based methods are used to assess \wtm. We test our system using two different LLMs, GPT-4-Turbo and GPT-4o-Mini, to compare their performance in story coherence, diversity, map enjoyment, objective enjoyment, aesthetic and functionality. Results indicate that GPT-4-Turbo outperforms GPT-4o-Mini in story coherence, diversity, objective enjoyment, and functionality. However, GPT-4o-Mini generates levels with greater aesthetic appeal, likely due to its tendency to use a more visually diverse selection of blocks. Both models perform similarly in map enjoyment.

The remainder of this paper is organized as follows. Section II reviews related work. Section III details the \wtm system, covering structured story generation with predefined elements (III-A), the main map generation process (III-B and III-C), and sub-map design (III-D). Section IV outlines the experimental setup, while Section V presents the results. Finally, Section VI concludes the paper and discusses directions for future work.


\section{Related Work}

As a system that starts from a prompt, extracts all relevant information, and generates a playable game in Minecraft, \wtm builds on prior research in generating games from stories. Ammanabrolu et al. \cite{ammanabrolu2020bringing} employ knowledge graphs to determine object placement, while Fan et al. \cite{fan2020generating} train a neural network on the Light dataset \cite{urbanek2019learning} to position objects within a game. Hartsook et al. \cite{hartsook2011toward} apply search-based optimization \cite{togelius2011search} to construct a graph that dictates object placement based on story elements. Our work extends Word2World \cite{nasir2024word2world}, a system that generates a story using an LLM and extracts the necessary information to create simplistic 2D role-playing games. We advance this approach by adapting the generated game into a complex 3D Minecraft environment, addressing limitations in Word2World through a scaling algorithm (discussed in Section \ref{sec:methodology}) and introducing sub-map generation to better handle different objectives.

\begin{figure*}[htb!]
    \centering
    \includegraphics[width=0.9\textwidth]{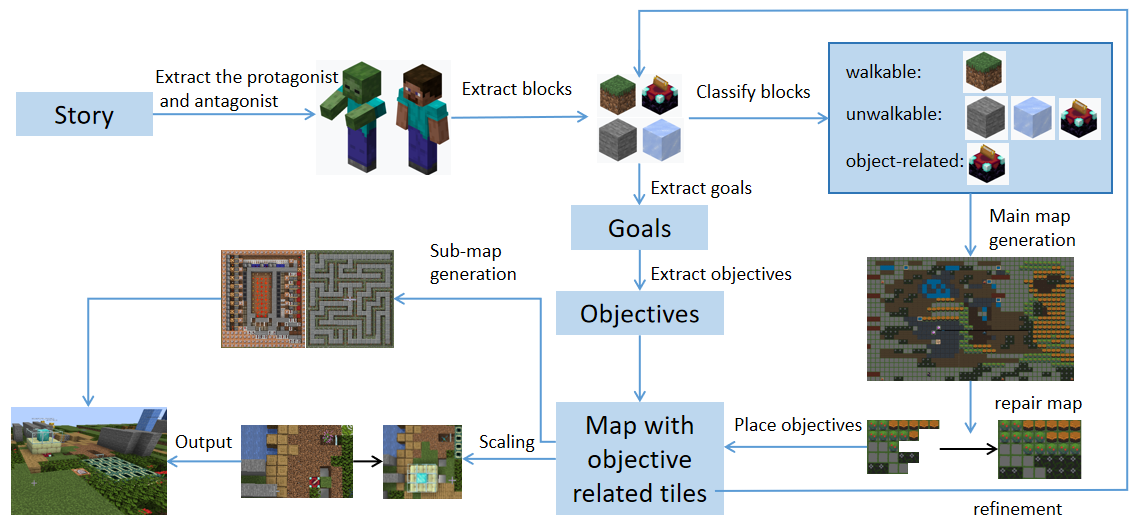}
    \caption{A flowchart depicting the complete \wtm pipeline. The LLM is used at every stage of the process.}
    \label{main_framework}
\end{figure*}

\wtm can be viewed as a system for procedural content generation (PCG)~\cite{shaker2016procedural}, a field that has seen significant advancements through deep learning and evolutionary algorithms~\cite{togelius2011search, summerville2018procedural, liu2021deep}. 
In previous research, level generation has often been treated as a sequence generation problem, leading to the use of sequence-based models like LSTMs~\cite{summerville2016super}. Building on this, transformer-based architectures \cite{vaswani2017attention} have demonstrated strong potential in PCG. For instance, Todd et al.~\cite{todd2023level} utilize GPT-2 and GPT-3 for Sokoban~\cite{murase1996automatic} level generation through supervised learning, while Sudhakaran et al.~\cite{sudhakaran2024mariogpt} fine-tune GPT-2 to generate Mario Bros. \cite{togelius2011procedural} levels, incorporating novelty search \cite{lehman2011novelty} for diversity.
Additionally, Nasir and Togelius~\cite{nasir2023practical} show that GPT-3 can generate novel levels using human-in-the-loop and bootstrapping methods~\cite{torrado2020bootstrapping} with minimal human-designed data. Minecraft has also been explored as a 3D PCG environment for LLMs and text-guided generation~\cite{awiszus2022wor, beukman2023hierarchically, earle2024dreamcraft}.

Beyond PCG, \wtm falls under the broader scope of automated game generation (AGG)~\cite{nelson2007towards}. 
Other AGG approaches include ANGELINA~\cite{cook2011multi, cook2012angelina}, which uses evolutionary algorithms to create puzzle and platformer games, and Genie/Genie-2~\cite{bruce2024genie, parkerholder2024genie2}, which trains large models on internet videos to generate platformer and 3D games from text prompts. \wtm also aligns with computational game creativity~\cite{browne2012computational, liapis2014computational}, as each stage of game generation---storytelling, character and tile description, and world construction---requires creative decision-making.

\section{Methodology}\label{sec:methodology}

This section introduces \wtm, a system that transforms stories into playable Minecraft levels. Each level consists of a main map and multiple sub-maps, generated through a step-by-step process. The following discussion details each stage of this process, with Figure~\ref{main_framework} providing an overview of the framework.


\subsection{Story generation using LLMs}
The level generation process begins with story creation. Since LLM-generated stories can vary widely, we impose constraints to ensure they contain the necessary elements for our games. Each story must feature a protagonist pursuing a goal, an antagonist attempting to thwart them, and additional NPCs if required. It should also establish a clear environment and include multiple objectives for the protagonist, one of which must involve defeating the antagonist. In our system, users can customize both the number of story paragraphs and the number of objectives, providing greater control over story complexity.

\subsection{Initial map generation}
The main game map serves as the foundation for each level. We use the LLM to generate a 2D layout of this map, ensuring it aligns with the narrative environment and objectives. Each tile in the 2D layout represents general objects or terrain features (e.g., trees or rivers), rather than specific Minecraft blocks. In the subsequent step, the LLM translates these general tiles into corresponding Minecraft blocks.

The LLM extracts key elements from the story---including basic information about the protagonist and antagonist, required tiles, walkable tiles, and objectives---to generate the tile-based main map layout. Additionally, it creates a tile-character mapping for the next step, assigning a unique character to each tile type.
Based on this information, the LLM generates an initial world layout represented as a tile-based grid. It then positions the protagonist, antagonist, and interactive objects on the map, assigning each objective to its corresponding location. Finally, the map is padded to form a rectangle.

After generating the world, both the LLM and traditional evaluation methods are used to assess its navigability, balance, and overall quality. Walkable paths are evaluated using the A* algorithm \cite{hart1968formal} to ensure that all objectives are reachable. During refinement, previous iterations of generated maps serve as references, enabling the model to learn from past designs.

\subsection{Adaptive tile scaling for enhanced realism}

The initial map generation suffers from a fatal scaling issue that significantly limits realism: each element, regardless of its real-world dimensions, is represented by a single tile. 
This limitation affects both aesthetics and realism, as large objects such as houses or mountains are unnaturally compressed. To address this challenge, we introduce a scaling algorithm that selectively expands certain tiles beyond their original size, enhancing both visual realism and the integrity of the game world.

\begin{figure}[htb]
    \centering

    \subfigure[Tile type array]{
	\label{compare0.sub.1}
	\includegraphics[width=0.53\linewidth]{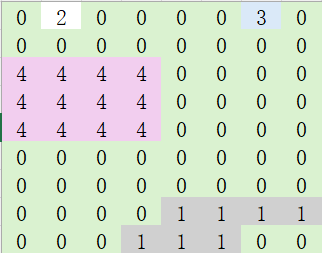}}
	\quad 
	\subfigure[Corresponding Minecraft level]{
	\label{compare0.sub.2}
	\includegraphics[width=0.37\linewidth]{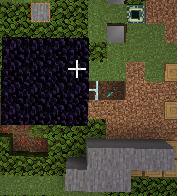}}
    
    \caption{Tile type array representation: 0 represents walkable tiles, 1 represents unwalkable tiles, 2 represents objective-related tiles, 3 represents tiles need to be scaled, and 4 represents already scaled tiles. This map will be updated by Algorithm \ref{scaling_ag}.}
    \label{tile_type_array}
\end{figure}

The first step in this process involves using the LLM to analyze the story and main map, identifying which tiles should be expanded and determining their target sizes. Each tile is categorized as walkable, unwalkable, objective-related, or already scaled, as depicted in Figure~\ref{tile_type_array}. Next, we apply Algorithm~\ref{scaling_ag} to select the highest-scoring valid tile for expansion. A tile is considered valid if scaling it would not overlap objective-related tiles or previously expanded tiles. The score for each candidate tile is calculated based on the summed occurrence frequencies of tiles it would cover in the original map, following the principle of ``rarity adds value'': covering scarce tiles could disrupt the environment coherence, whereas overlapping common tiles like grass minimally impacts the overall map integrity.


\begin{figure}[htb]
    \centering
    \includegraphics[width=.48\textwidth]{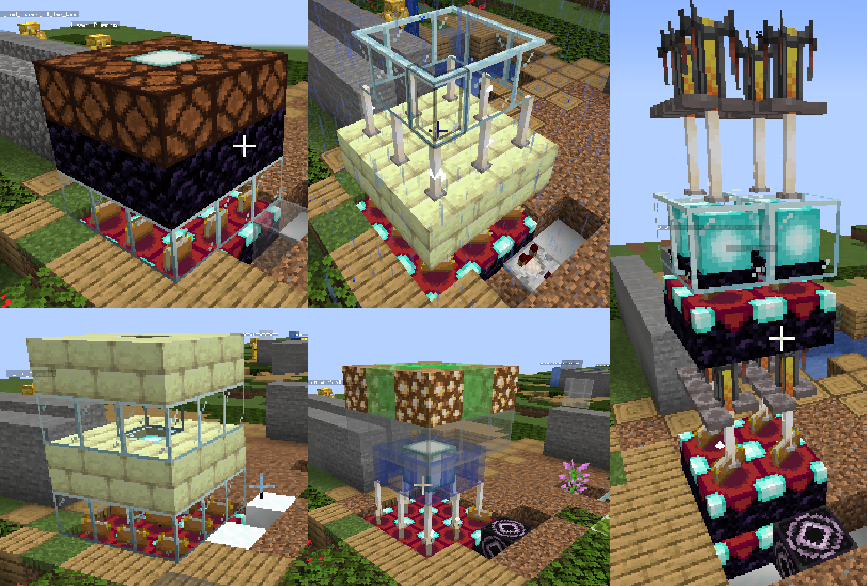}
    \caption{LLM-generated buildings, each expanded from a tile labelled ``Illusory Object Tile.''}
    \label{buildings}
\end{figure}

These expanded tiles are then replaced with larger structures generated by the LLM, taking into account the story context and the descriptions of the original tiles.
We note that a key property of the LLM is its ability to introduce diversity; even for identical tile types, it produces stylistically consistent yet visually distinct buildings, as illustrated in Figure~\ref{buildings}.

\begin{algorithm}[htbp]
\caption{Scaling algorithm}\label{scaling_ag}
\KwData{Main map layout $M$, List of tiles requiring scaling $C$, Dictionary of scaling sizes \( D = \{ c_i \mapsto s_i \} \), where \( c_i \) is a tile character and \( s_i \) is its scaling size}

\For{i = 1 to $|M|$}{
    \For{j = 1 to $|M[0]|$}{
        \If{$M[i][j]$ in $C$}{ 
            $t \leftarrow M[i][j]$ \tcp{Target tile}
            $s \leftarrow D[t]$ \tcp{scaling size}
            $best\_m \leftarrow -1$, $best\_n \leftarrow -1$ \tcp{Best top-left corner}
            $best\_score \leftarrow 0$\\
            \For{$m = i - s + 1$ to $i$}{
                \For{$n = j - s + 1$ to $j$}{
                    $score \leftarrow$ calculate score for $s \times s$ block with $(m, n)$ as top-left corner
                    \If{$score > best\_score$}{
                        $best\_score \leftarrow score$\\
                        $best\_m \leftarrow m$\\
                        $best\_n \leftarrow n$
                    }
                }
            }
            \If{$best\_m!=-1$}{
                update the map
            }
        }
    }
}
\end{algorithm}

\subsection{Sub-map generation}

\begin{figure}[htbp] 
	\centering  
    
	\subfigure[Learn Old Farming Techniques from Señora Miro (``Chat with NPC'')]{
	\label{compare1.sub.1}
	\includegraphics[width=0.25\linewidth]{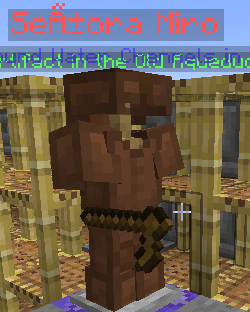}}
	\quad 
	\subfigure[Navigate through the Sand Serpent Dunes (``Exit Maze'')]{
	\label{compare1.sub.2}
	\includegraphics[width=0.35\linewidth]{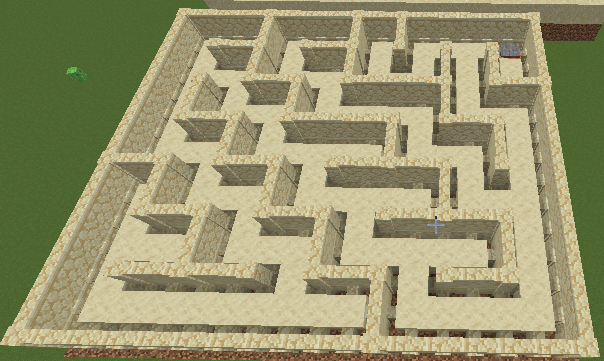}}
	\subfigure[Collect samples from the Cactus Gates (``Collect Items'')]{
	\label{compare1.sub.3}
	\includegraphics[width=0.25\linewidth]{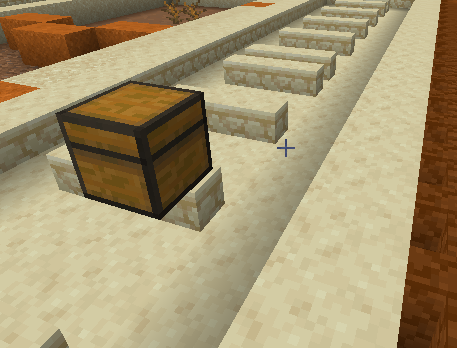}}
    
	\quad
	\subfigure[Dodge attacks from local wildlife (``Survive Waves of Enemies'')]{
	\label{compare1.sub.4}
	\includegraphics[width=0.35\linewidth]{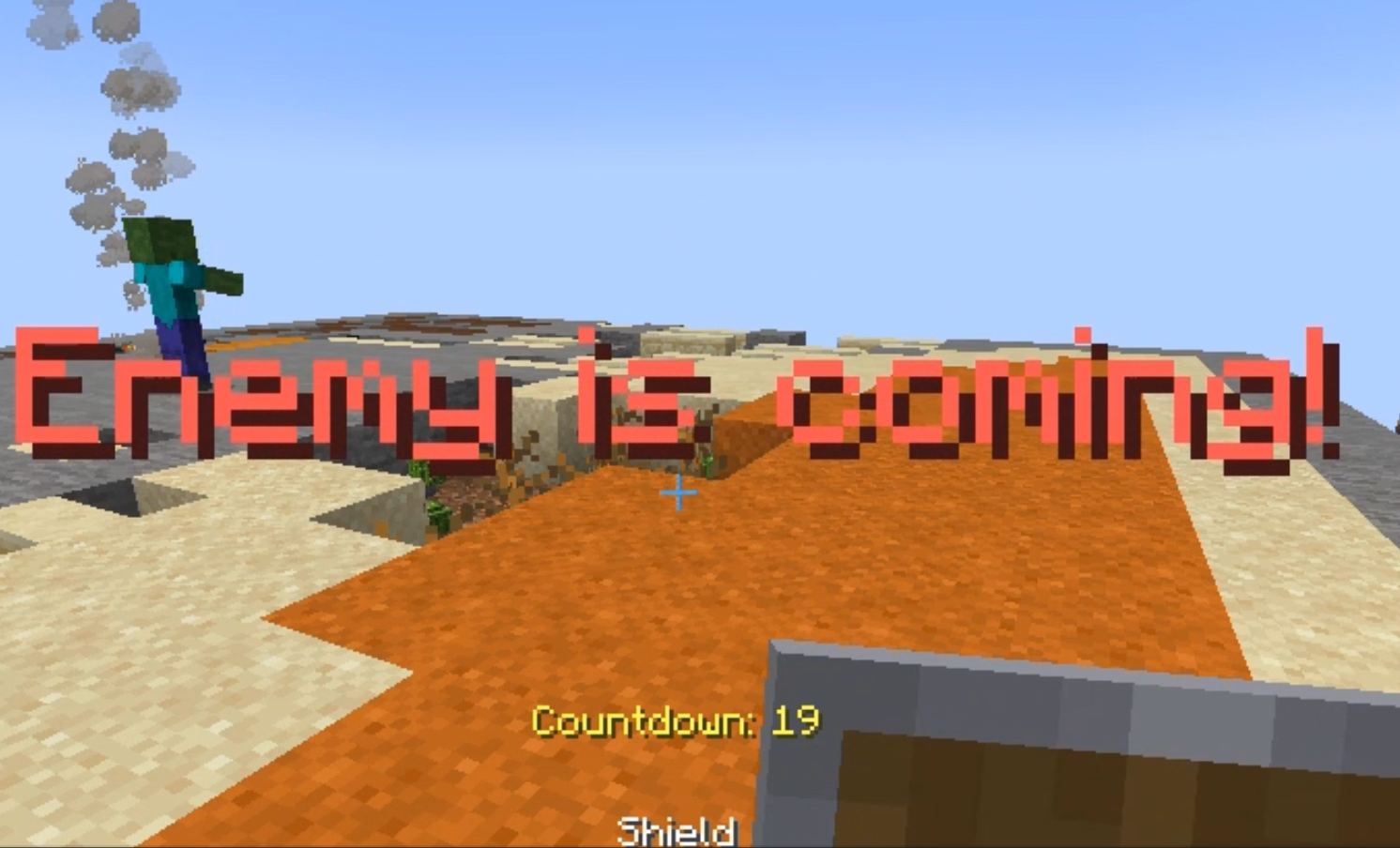}}
	\subfigure[Defeat Vorath to secure the Bloom of Night (``Defeat the Enemy'')]{
		\label{compare1.sub.5}
		\includegraphics[width=0.5\linewidth]{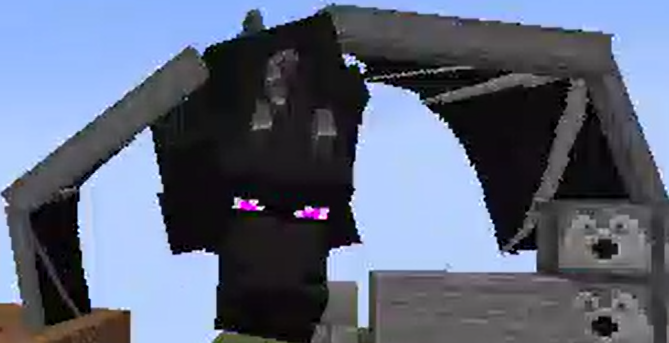}}
	\caption{Examples of objective-oriented quests.}
	\label{objective}
\end{figure}

Prior work \cite{aarseth2005hunt} categorizes game quests into three fundamental types: place-oriented, time-oriented, and objective-oriented. We further divide objective-oriented quests into five specific types: ``Defeat the Enemy'', ``Chat with NPC'', ``Exit Maze'', ``Survive Waves of Enemies'', and ``Collect Items'', as illustrated in Figure~\ref{objective}.


However, it is challenging to directly implement all objectives within the main map due to constraints in size and complexity. For instance, maze navigation requires structured layouts incompatible with the open-world style of the primary map. To address this, we introduce tailored sub-maps specifically for the ``Exit Maze'', ``Survive Waves of Enemies'', and ``Collect Items'' objectives using the following process:

\begin{enumerate}
    \item \textbf{Objective Position Generation}: We prompt the LLM to identify a representative tile and determine its position on the main map for each objective.

    \item \textbf{Position Adjustment}: Since LLM-generated coordinates may sometimes be invalid (e.g., placed within obstacles or mismatched tile types), we employ a BFS algorithm to locate the nearest valid position, and place a portal at that location instead.

    \item \textbf{Sub-map Generation}: We generate structured sub-maps tailored to specific objectives. These sub-maps connect directly to the main map via the previously placed portals.
\end{enumerate}

\section{Experiments}

For metric-based evaluation, we conduct experiments using two language models---GPT-4-Turbo and GPT-4o-Mini---to evaluate \wtm. Each model generates 30 game levels, which are then assessed based on story coherence, diversity, enjoyment, aesthetics, and functionality. For the enjoyment metric, we additionally compare these results against those produced by an evolutionary algorithm as a baseline.

We adopt hyperparameters similar to those used by Nasir et al.~\cite{nasir2024word2world}, configuring the game levels with a story comprising 4–5 paragraphs and assigning the protagonist 8 objectives to complete each game. The A* algorithm is limited to 1000 iterations when searching for paths to all objectives in the initial map.

For human-based evaluation, we conduct a user study to further assess GPT-4-Turbo's performance regarding story coherence, visual appearance, and functionality. A total of 17 participants with basic knowledge of Minecraft and Game AI took part in this study.

\subsection{Metric-based story coherence evaluation}
Story coherence measures how well the generated levels match their original stories. In the metric-based story coherence evaluation, we employ both direct coherence evaluation and reconstructed story-based similarity analysis.

\subsubsection{Direct coherence evaluation}
 We provide the LLM with the original story, the tile-character mapping, and the character-based tile encoding of the level’s main map. The model is then prompted to assess the coherence between the generated map and the story.

\subsubsection{Reconstructed story-based similarity analysis}
\begin{figure}[htb]
    \centering
    \vspace{-0.1cm}
    \includegraphics[width=0.45\textwidth]{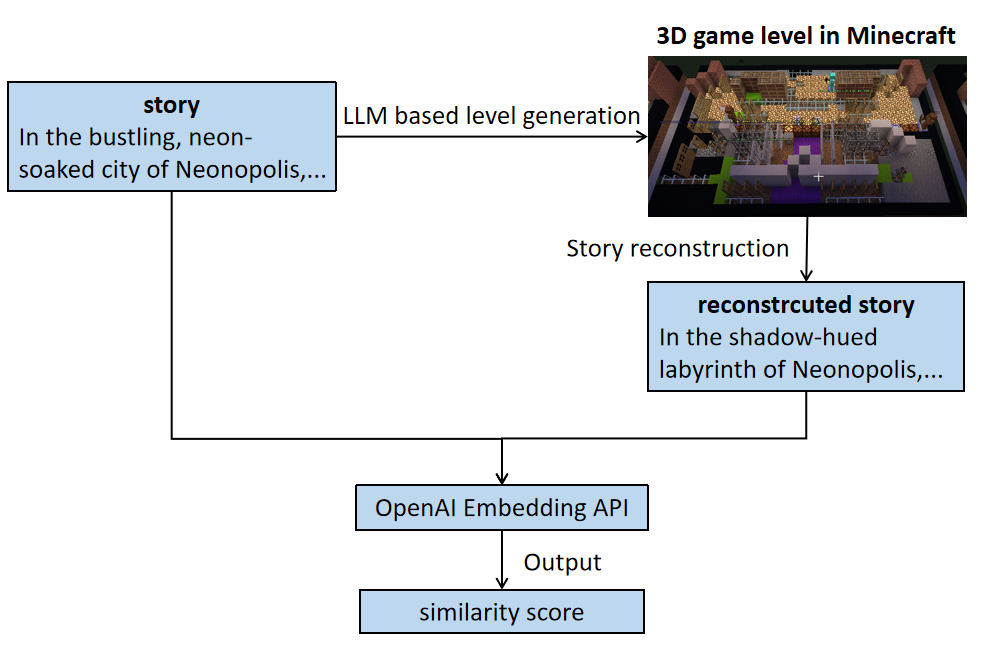}
    \caption{Reconstructed story-based similarity analysis}
    \label{recon_story_eva}
\end{figure}

We extract the block-level representation of the level generated within Minecraft as a JSON file. Next, we prompt the LLM to generate a new story based on these extracted blocks. To quantify coherence, we compute the cosine similarity between the original and reconstructed stories using \texttt{text-embedding-3-large} \cite{wang2023improving}, a high-dimensional embedding model that captures semantic closeness as shown in Figure~\ref{recon_story_eva}.

Table \ref{metrics_story_coherence} shows that GPT-4-Turbo has higher average coherence score and lower standard deviation, which implies that it is more stable and performs better in maintaining story coherence.

\begin{table}[htb]
    \centering
    \vspace{-0.2cm}
    \caption{Metric-based story coherence evaluation (Mean \& Standard Deviation)\label{metrics_story_coherence}}
    \begin{tabular}{cccccc}
        \toprule
        &{Model} & {Direct Coherence} & {Reconstructed Story Coherence}\\
        \midrule
        \multirow{8}{*}{}
        & GPT-4-Turbo & 84.47 $\pm$ 8.87 & 0.90 $\pm$ 0.01 \\
        & GPT-4o-Mini & 71.75 $\pm$ 13.99 & 0.84 $\pm$ 0.04 \\
        \bottomrule
    \end{tabular}
\end{table}

 

\subsection{Human-based story coherence evaluation}

\subsubsection{User study for story coherence of the main map}

We designed two questions to evaluate GPT-4-Turbo’s ability to generate levels consistent with given stories. In each question, participants were presented with a short story and two top-down images of Minecraft levels: one generated from the provided story and another from a different story with a similar theme (e.g., both set in a forest). Participants were asked to identify which level matched the given story. This design specifically ensures that participants cannot rely solely on visual cues, as both maps share similar thematic elements, encouraging them to consider the deeper narrative and structural coherence instead. The accuracy of participants’ choices shows how well our generated levels maintain coherence with the original story.


\subsubsection{User study for story coherence of the sub-map}
When evaluating the coherence between a sub-map and its corresponding objective, we offer participants the description of the objective, relevant part in the story, and two sub-maps. Both sub-maps are generated from different objectives of this story, whose similar block compositions make this task challenging. Therefore, a high accuracy in this task would further indicate a strong alignment between the sub-map and its objective.

\subsubsection{User study for level similarity}

We maintain that the levels generated from the same story should follow a consistent distribution in terms of style and structure. In this experiment, each participant is presented with a base level and three candidate levels, among which only one level is generated from the same story as the base level.

The results are shown in Table \ref{user_story_coherence}. In both themes, sub-maps achieve higher coherence scores (0.82) compared to main maps (0.59 for forest and 0.47 for desert). A possible explanation is that main maps are designed to represent the entire story, which contains fewer details. In contrast, sub-maps focus on a specific objective, and participants only need to read a small portion of the story rather than processing the entire narrative.

We also find that the desert theme main map shows lower coherence than the forest one. This is due to the limited block variety in desert maps, where sand and stone dominate both answer choices, making it difficult to distinguish. Additionally, the correct map was smaller and structurally simpler compared to the distractor, which may also lead to the low accuracy in this map.

Overall, the level similarity accuracy is moderate (0.59), likely because perceived similarities in top-down views are influenced more by overall visual themes rather than specific narrative coherence.

\begin{table}[htb]
    \centering
    \caption{Human-based story coherence evaluation\label{user_story_coherence}}
    \begin{tabular}{cccccc}
        \toprule
        &{Question} & {Accuracy}\\
        \midrule
        \multirow{8}{*}{}
        & Forest theme main map & 0.59\\
        & Forest theme sub map & 0.82\\
        & Desert theme main map & 0.47\\
        & Desert theme sub map & 0.82\\
        & Level similarity & 0.59\\
        \bottomrule
    \end{tabular}
\end{table}

\subsection{Diversity}
Diversity is a crucial aspect of game-level design~\cite{li2025measuring}, as it enriches gameplay by providing players with varied experiences.
In our experiments, we evaluate diversity through tile variety, a key metric that measures environmental richness. Specifically, we quantify tile variety using Shannon Entropy, defined as follows. Given a map containing $N \in \mathbb{N}_+$ distinct tile types, the Shannon Entropy ($H$) is calculated as shown in Equation (1), where $p_i$ represents the proportion of the $i^{th}$ tile type:
\vspace{-0.25cm}
\begin{eqnarray}
& & H = - \sum_{i=1}^{N} p_i \log_2 p_i.
\end{eqnarray}

Results from our evaluation, shown in Table~\ref{metrics_tile_variety}, indicate that maps generated by GPT-4-Turbo exhibit greater diversity, reflected by higher tile variety and Shannon entropy, compared to those created by GPT-4o-Mini. This demonstrates that GPT-4-Turbo generates more varied and visually interesting game environments.


\begin{table}[htb]
    \centering
    \vspace{-0.3cm}
    \caption{Metric-based tile variety evaluation (Mean \& Standard Deviation)\label{metrics_tile_variety}}
    \begin{tabular}{cccccc}
        \toprule
        &{Metric} & {GPT-4-Turbo} & {GPT-4o-Mini}\\
        \midrule
        \multirow{8}{*}{}
        & Tile Type Number & 18.23 $\pm$ 5.70 & 12.95 $\pm$ 2.04 \\
        & Shannon Entropy & 4.06 $\pm$ 0.49 & 3.58 $\pm$ 0.33 \\
        \bottomrule
    \end{tabular}
\end{table}

 


\subsection{Enjoyment}
We measure the enjoyment of the levels from two aspects: map enjoyment and objective enjoyment. We also employ an evolutionary algorithm (EA) as the baseline to compare the LLM's performance with traditional PCG. For a fair comparison, the chromosome in the EA is set to a $15 \times 15$ grid-based map---the smallest size among the LLM-generated maps. The fitness is the average of the shortest paths from the starting position to all objective tiles (ASPAO)

\subsubsection{Map enjoyment}
Playability is the cornerstone of map enjoyment in any game. Although the A* algorithm is used to ensure that the protagonist can access all objectives, scaling can block some original paths. We applied a BFS check after scaling to verify that all objective-related tiles on the map remain connected by at least one valid path.

As long as the map remains valid, we assume that a high proportion of unwalkable tiles relative to the total area indicates high map enjoyment, as a greater density of unwalkable tiles forces players to strategize their movement and navigate dynamic obstacles. Therefore, we use the valid unwalkable tiles ratio (VUTR) instead of the unwalkable tiles ratio (UTR) to represent map enjoyment.

We define $\mathcal{A}_m$ as the area of map $m$ and $\mathcal{U}_m$ as the area of unwalkable tiles in the map $m$. Assuming we have $M$ maps, $UTR$ and $VUTR$ can be modeled as (2) and (3), where $C_m \in \{0,1\}$ denotes the validity of map $m$, with $C_m=1$ if map $m$ is valid.
\vspace{-0.1cm}
\begin{eqnarray}
& & UTR = \sum_{m=1}^{M} \frac{ \mathcal{U}_m}{\mathcal{A}_m}\\
& & VUTR = \frac{\sum_{m=1}^{M} C_m \cdot \mathcal{U}_m}{\sum_{m=1}^{M}\mathcal{A}_m}
\end{eqnarray}



\subsubsection{Objective enjoyment}
While a high proportion of unwalkable tiles generally represent enjoyment, this metric may fail in certain extreme cases. For instance, if a large number of unwalkable tiles are set at the edges of the map while the objectives are concentrated at the center, the actual challenge for players may remain low.

The average of the shortest paths from the starting position to all objective tiles (ASPAO) among valid maps is a key metric for evaluating the objective enjoyment of a game level, as the player is teleported back to the starting position after completing an objective. A higher average distance suggests a more complex level, requiring players to traverse longer distances to reach objectives.  ASPAO can be modeled in (4) and (5), where $d(s,o)$ is the shortest distance from the starting point $s$ to objective tile $o$, and $P(s,o)$ is the set of shortest paths.
\vspace{-0.1cm}
\begin{equation}
ASPAO = \frac{1}{|O|} \sum_{o \in O} d(s, o)
\end{equation}
\begin{equation}
d(s, o) = \min_{\pi \in P(s, o)} |\pi|
\end{equation}

We compare the enjoyment levels of the two LLM models and EA, with the results presented in Table \ref{metrics_enjoyment}. We observe that EA achieves the best performance across all metrics, which is expected since it optimizes solely for ASPAO rather than trying to make a balance between ASPAO and other metrics like story coherence. This direct optimization also improves the final VUTR by increasing the ratio of unwalkable tiles, thereby enhancing level playability. These findings highlight the significant gap between LLM and EA in terms of enjoyment.

Additionally, we find that GPT-4o-Mini achieves a higher playability score (0.87) compared to GPT-4-Turbo (0.82) while maintaining a lower UTR. This suggests that GPT-4-Turbo adopts a more aggressive scaling strategy when generating levels, potentially blocking important paths and increasing the proportion of unreachable objectives. As a result, the VUTR remains similar between the two models, indicating comparable map enjoyment. The higher ASPAO of GPT-4-Turbo further suggests that its generated objectives are more engaging.


\begin{table}[htb]
    \vspace{-0.2cm}
    \centering
    \caption{Metric-based enjoyment evaluation (Mean \& Standard Deviation)\label{metrics_enjoyment}}
    \begin{tabular}{cccccc}
        \toprule
        &{Metric}& {GPT-4-Turbo} & {GPT-4o-Mini}&{EA}\\
        \midrule
        \multirow{4}{*}{}
        & Playability & 0.83 $\pm$ 0.37 & 0.87 $\pm$ 0.35 & \textbf{1 $\pm$ 0}\\
        & UTR & 0.18 $\pm$ 0.16 & 0.14 $\pm$ 0.12& \textbf{0.40 $\pm$ 0.02} \\
        & VUTR & 0.11 $\pm$ 0.13 & 0.12 $\pm$ 0.11& \textbf{0.40 $\pm$ 0.02} \\
        & ASPAO & 22.41 $\pm$ 9.28 & 10.05 $\pm$ 5.66 & \textbf{33.4 $\pm$ 1.85} \\
        \bottomrule
    \end{tabular}
    \vspace{-0.2cm}
\end{table}


\subsection{Aesthetic and functionality}
To evaluate the aesthetic and functionality of levels generated by GPT-4-Turbo, we use GPT-4o-Mini as a baseline. We record a 2-3 minute gameplay video showcasing a generated level for each model. Participants are then asked to compare the two videos and rank the levels based on aesthetic appeal and functionality separately.

The results in Table \ref{user_aesthetic} show that GPT-4-Turbo outperforms GPT-4o-Mini in functionality. However, participants perceive GPT-4o-Mini as having better aesthetics. We believe this may be due to GPT-4o-Mini selecting blocks with weaker story coherence, leading to more contrasting block choices. For example, in a forest-themed level, GPT-4o-Mini incorporated lava blocks, resulting in a more colorful and visually striking map, which could appeal to participants’ aesthetic preferences.

\begin{table}[htb]
    \vspace{-0.2cm}
    \centering
    \caption{Human-based Aesthetic and functionality evaluation\label{user_aesthetic}}
    \begin{tabular}{cccccc}
        \toprule
        &{Metric}& {GPT-4-Turbo} & {GPT-4o-Mini}\\
        \midrule
        \multirow{4}{*}{}
        & Aesthetic & 4 & \textbf{13} \\
        & Functionality & \textbf{11} & 6\\
        \bottomrule
    \end{tabular}
    \vspace{-0.2cm}
\end{table}

\begin{figure}[htbp] 
	\centering  
    
	\subfigure[Map A: GPT-4-Turbo with scaling]{
	\label{compare2.sub.1}
	\includegraphics[width=0.82\linewidth]{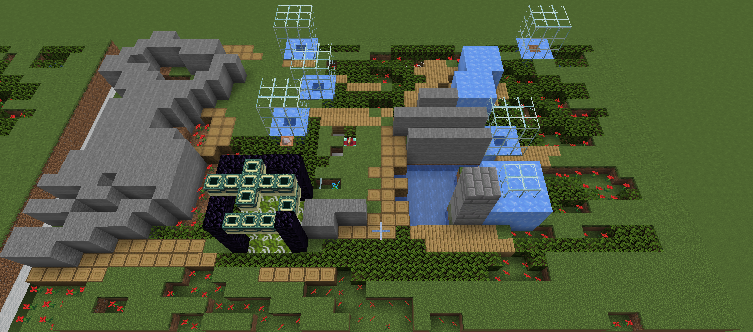}}
    
	\subfigure[Map B: GPT-4-Turbo without scaling]{
	\label{compare2.sub.2}
	\includegraphics[width=0.82\linewidth]{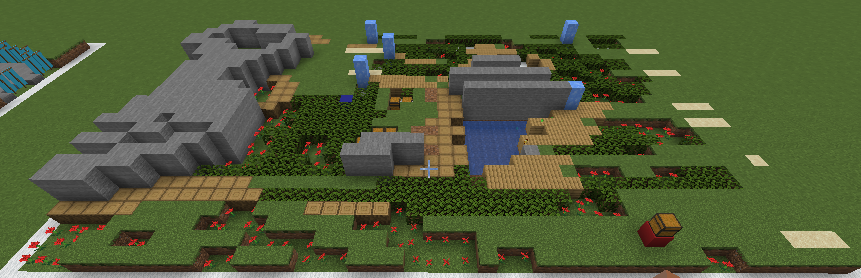}}
    
	\subfigure[Map C: GPT-4o-Mini with scaling]{
	\label{compare2.sub.3}
	\includegraphics[width=0.82\linewidth]{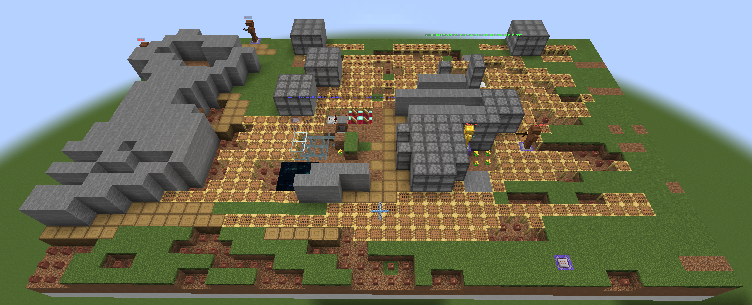}}
    
	\subfigure[Map D: GPT-4o-Mini without scaling]{
	\label{compare2.sub.4}
	\includegraphics[width=0.82\linewidth]{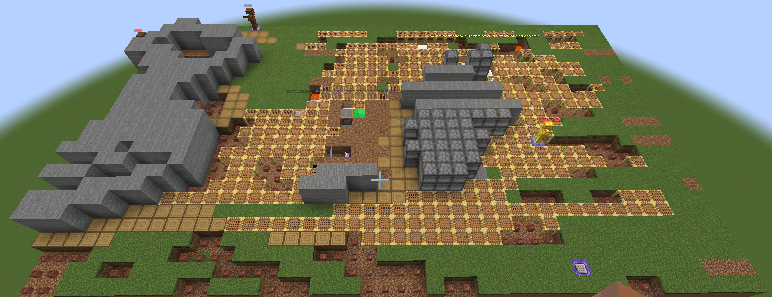}}
	\caption{Ablation study}
	\label{ablation study}
    \vspace{-0.2cm}
\end{figure}

\subsection{Ablation Study}
We conduct an ablation study to determine whether the scaling algorithm or the LLM's block selection plays a more critical role in the post-processing phase after generating the initial map. We prepare four levels that share the same initial main map but differ in their post-processing methods: A (GPT-4-Turbo with scaling), B (GPT-4-Turbo without scaling), C (GPT-4o-Mini with scaling), and D (GPT-4o-Mini without scaling), as shown in Figure~\ref{ablation study}. Participants are asked to rank these levels based on their preference. The composite score $S$ is calculated as a weighted average of the votes across different rankings, providing an overall assessment of the results. Given $k$ ranking categories, where each rank $i$ has a corresponding vote count $R_i$ and weight $w_i$, the composite score $S$ can be expressed as (6), where $\mathcal{N}:= \sum_{i=1}^{k} R_i$ is the total number of votes.
We set $k=4$ and $w_i=k-i$ in our experiments.
\vspace{-0.1cm}
\begin{equation}
S = \frac{\sum_{i=1}^{k} w_i R_i}{\mathcal{N}}
\end{equation}

As shown in Table~\ref{ablation study res}, Map A achieves the best performance. Additionally, we observe that Map B has a higher composite score than Map C, indicating that the LLM's block selection has a greater impact on the final quality of post-processed maps than the scaling algorithm.

\begin{table}[htb]
    \vspace{-0.2cm}
    \centering
    \caption{Ablation study results\label{ablation study res}}
    \begin{tabular}{cccccc}
        \toprule
        &{Map}& {Composite score}\\
        \midrule
        \multirow{4}{*}{}
        & Map A & \textbf{3.24} \\
        & Map B & 2.76 \\
        & Map C & 2.29 \\
        & Map D & 1.35 \\
        \bottomrule
    \end{tabular}
\end{table}


\section{Conclusion and future work}

In this paper, we introduce \wtm, a novel system that utilizes LLMs to generate structured and playable Minecraft levels from stories. Our approach addresses key challenges in story-driven PCG by incorporating a scaling algorithm to preserve spatial consistency and sub-map generation to support diverse objectives. Through both metric-based and human-based evaluations, we found that GPT-4-Turbo produces levels with higher story coherence, diversity, and functionality, while GPT-4o-Mini excels in aesthetic appeal. Our ablation study further highlighted the importance of block selection by the LLM in determining the final quality of generated levels.

In the future, we aim to improve level integration by merging all objectives into a single continuous map, rather than relying on portals to connect sub-maps. This will require a more advanced scaling algorithm. 
Additionally, our current maps assume uniform height for all walkable tiles, which does not fully utilize Minecraft's action space, including jumping and climbing mechanics. We plan to introduce more complex terrain variations to enhance verticality and traversal dynamics, making levels more engaging and challenging.

Our work demonstrates the potential of LLMs in procedural content generation, particularly in translating narratives into structured, playable game environments. By further exploring hybrid approaches that combine LLMs with traditional PCG techniques, we aim to push the boundaries of story-driven level generation, making AI-generated game worlds more immersive, coherent, and engaging for players.

\bibliographystyle{IEEEtran} 
\bibliography{minecraft}

\section{Appendix: Important LLM prompt}
\subsubsection{Story generation}
Write a 4-5 paragraph story which has characters including the protagonist trying to achieve something and the antagonist wanting to stop the protagonist. There should be 8 objectives for the protagonist in the story. One of them should be to defeat the antagonist somehow. You should describe the environments that those objectives happen. You can add some NPCs in the story.
\subsubsection{Character extraction}
Let's use the above story to create a 2D game. Write a specific description of each character which can be used as a prompt to generate sprites for the characters.
\subsubsection{Tile extraction}
Create an exhaustive list of tiles needed to create the environment. Some tile can occupy more than one space.

\subsubsection{Tile-character mapping generation}
Imagine each tile maps to an alphabet or a character. For environment, use alphabets and for characters use special characters. Create it in a single Python Dictionary style. Return only and only a Python Dictionary and nothing else in your response. Don't return it in a Python response. Names should be the Keys and alphabets or characters should be the Values. Protagonist should always strictly be '@' and the antagonist should always strictly be '\#'.

\subsubsection{World generation}
 Using the following tile to character mapping:\{tile\_map\_dict\}. Create an entire world on a tile-based grid. Do not create things that would need more than one tile. Also, following characters are important to place:\{important\_tiles\_list\}, walkable tiles:\{walkable\_tiles\_list\}. Do not place the protagonist, the antagonist and the interactive objects of the story right now. Only create the world right now. Create it is a string format with three backticks to start and end with (```) and not in a list format."

\subsubsection{Objective Placement}
You are a great planner in 2D game. You plan objectives for the protagonist of the game. All objectives should match the goals extracted from the story. One of them should be to defeat the antagonist somehow. Other objectives can be: finding the exit of a complex labyrinth, finding a chest in the map, surviving waves of enemies, or gathering some items in the map. Return a Python dictionary of the objective as the key and a tile that achieves the objective and the position of the tile. For example `Objective': ['A', 6, 1]. Only return a Python dictionary. Do not return a python response.
\subsubsection{Scaling tiles selection}
Given the story, a 2D map \{tile\_map\}, and a dictionary {des2not} where each key is a tile\'s description and each value is the notation for that tile in the map, identify which tile notations in the map need to be scaled? A tile is considered scaled if it should occupy more than one grid cell, such as a \'house\' tile. \# Please avoid selecting adjacent tiles of the same type and frequent tiles. Please avoid scaling \# and @ Return a Python list of tiles, formatted as a list of scaled tile notations (e.g., [a,b]])'
\end{document}